\documentclass{article}

\PassOptionsToPackage{numbers, compress}{natbib}
\usepackage{lineno,hyperref}
\usepackage{float}
\usepackage[final]{neurips_2022}
\usepackage{bbm}
\usepackage{graphicx}
\usepackage{subfigure}
\usepackage[utf8]{inputenc} 
\usepackage[T1]{fontenc}    
\usepackage{hyperref}       
\usepackage{url}            
\usepackage{booktabs}       
\usepackage{amsfonts}       
\usepackage{nicefrac}       
\usepackage{microtype}      
\usepackage{xcolor}         
\usepackage{amsmath}
\usepackage{authblk}
\usepackage{tikz}
\usetikzlibrary{calc}
\usetikzlibrary{shapes}
\usepackage{adjustbox}
\usepackage{bm}
\usepackage{multirow}
\usepackage{blindtext}
\usepackage{algorithm}
\usepackage[noend]{algpseudocode}
\usepackage{pgfplots}
\usepackage{float}
\usetikzlibrary{calc,positioning,shapes,fit}
\pgfplotsset{compat=1.17} 

\tikzset{pblock/.style = {rectangle split,
                      rectangle split parts=2, very thick,draw=black!50, top
                      color=white,bottom color=black!20, align=center}}
 
 \tikzset{ldgrayblock/.style = {rectangle split,
                      rectangle split parts=2, very thick,draw,rectangle split part fill={white,black!20}, align=center}}

\tikzset{twoblock/.style = {rectangle split,
                      rectangle split parts=2, very thick,draw, align=center}}        
                      
\tikzset{longtwoblock/.style = {rectangle split,
                      rectangle split parts=2, very thick, align=center, draw, text width=12cm}}                       
                      
\tikzstyle{process} = [rectangle, minimum width=4cm, minimum height=1cm, inner sep=3mm, text centered,  very thick, align=center,draw=black]

\tikzstyle{crect} = [rectangle, rounded corners, minimum width=2cm, minimum height=1cm, inner sep=3mm, text centered,  very thick, align=center,draw=black]
 
\tikzstyle{decision} = [diamond, aspect=1, text centered,  very thick, align=center,  minimum width=1mm, minimum height=1mm, draw=black]

\tikzstyle{circmarker}=[circle,  text centered,  very thick, align=center,  minimum width=2mm,draw=black,fill=white,inner sep=0.5mm]

\tikzstyle{circstate}=[circle,  text centered,  very thick, align=center,  minimum width=1mm,draw=black,fill=white,inner sep=1mm]

\tikzstyle{circH}=[circle,  text centered,  very thick, align=center,  minimum width=3mm,draw=black,fill=white,inner sep=1mm]

 \tikzset{every fit/.style={text width=20cm, inner sep=3mm}}
\def\Temp{\mathrm{Temp}}

\title{Learning on Graphs for Mineral Asset Valuation Under Supply and Demand Uncertainty}

\author[1]{Yassine Yaakoubi}
\author[2]{Hager Radi}
\author[1]{Roussos Dimitrakopoulos}
\affil[1]{COSMO -- Stochastic Mine Planning Laboratory\\
McGill University, Quebec, Canada\\
}
\affil[2]{
Mila, Quebec AI Institute, Canada

\vspace{-2mm}
\texttt{\{yassine.yaakoubi,roussos.dimitrakopoulos\}@mcgill.ca} \\
\texttt{hager.radi@mila.quebec}
}

\begin{document}

\maketitle
\vspace{-3mm}
\vspace{-3mm}
\begin{abstract}
Valuing mineral assets is a challenging task that is highly dependent on the supply (geological) uncertainty surrounding resources and reserves, and the uncertainty of demand (commodity prices). In this work, a graph-based reasoning, modeling and solution approach is proposed to jointly address mineral asset valuation and mine plan scheduling and optimization under supply and demand uncertainty in the "mining complex" framework. Three graph-based solutions are proposed: (i) a neural branching policy that learns a block-sampling ore body representation, (ii) a guiding policy that learns to explore a heuristic selection tree, (iii) a hyper-heuristic that manages the value/supply chain optimization and dynamics modeled as a graph structure. Results on two large-scale industrial mining complexes show a reduction of up to three orders of magnitude in primal suboptimality, execution time, and number of iterations, and an increase of up to $40\%$ in the mineral asset value.
\end{abstract}
\vspace{-4mm}

\section{Introduction}
\vspace{-1mm}
Mineral asset valuation refers to the process of estimating the current market value of a mineral asset, using an established valuation method~\citep{jones2018}. It is a challenging task given the degree of geological uncertainty surrounding resources and reserves, and the uncertainty in commodity prices over the life of the mining asset. The value of a mining asset is typically forecast using the net present value (NPV) or discounted cash flow model. The standard practice is to generate a mine plan that considers the individual components separately and maximizes NPV in a deterministic manner, thus not accounting for supply (geological) uncertainty in any of the forecasting steps, and only accounting for demand (commodity price) uncertainty in post-processing. In this paper, a graph-based reasoning, modeling and solution approach is proposed to jointly address mineral asset valuation and mine plan scheduling and optimization under supply and demand uncertainty in the "mining complex" framework.

An industrial mining complex consists of a set of operations that manage the extraction of materials from several mines, their processing and flow through the mineral value/supply chain, to the generation of a set of sellable products and all intermediate components (production scheduling, blending, stockpiling, equipment capacities, processing streams, waste and tailings disposal, capital investments, etc.)~\citep{dimitrakopoulos2022,goodfellow2016}. All of these operations constitute a complex non-linear system that requires a strategic mine plan to maximize the NPV while meeting various complex constraints. Conventional mine planning methods break down optimization into distinct sequential steps and ignore geological uncertainty to reduce the model size and allow the use of commercial solvers. This is known to be the main cause of failing to meet production targets and forecasts~\citep{dimitrakopoulos2011}. To circumvent these limitations, the simultaneous stochastic optimization of mining complexes (SSOMC) integrates all interrelated components of the mineral value chain under geological uncertainty into a single formulation. To solve the SSOMC, a simulated annealing based solver was proposed and found to provide state-of-the-art results\citep{dimitrakopoulos2022, goodfellowThesis, goodfellow2016, saliba2019}. This solver is referred to in this manuscript as the baseline.

\vspace{-0.3cm}
\paragraph{Mine planning and optimization under commodity price uncertainty:}

Despite the large body of work on mine planning and optimization, very few studies~\citep{castillo2014,rimele2020,saliba2019} have addressed the incorporation of market uncertainty into mine planning and optimization~\citep{castillo2014,saliba2019} or SSOMC~\citep{saliba2019}. The price of the finished products naturally governs the profitability and feasibility of the operation; thus, ignoring price volatility will result in a suboptimal (misguided) mine plan.
The only reason that price uncertainty has not been considered to date is the additional computational complexity that would be added, making an already complex large-scale nonlinear stochastic industrial problem simply intractable.
Indeed, the combined uncertainty, and thus the size of the model, grows exponentially with the number of uncertainty sources.
To avoid these computational limitations, the standard practice has always been to assume a deterministic and constant commodity price over the life of the mine~\citep{dimitrakopoulos2022,goodfellow2016}. Past work has examined the possibility of including price uncertainty as a preprocessing step that precedes the mine planning and optimization phase~\citep{castillo2014} or a post-processing step that follows it~\citep{rimele2020}. Recent work looked at incorporating commodity price uncertainty into SSOMC, generating a more robust long-term plan in that the generated plan better manages and quantifies the risk derived from spot market volatility, even when using a limited number of commodity price realizations.

\vspace{-0.3cm}
\paragraph{Graph representation learning:} In this work, graph-based reasoning, modeling, and representation learning are used to develop a solution methodology that would produce a mineral asset valuation for real large-scale industrial mining complexes in conjunction with the schedule, rather than a post-processing step as in established methods. In recent years, there has been an increasing body of research looking into using graph-based reasoning to combine artificial intelligence (AI) and operations research (OR) for tackling combinatorial optimization problems~\citep{Bello2017, gasse2019, morabit2021, Yaakoubi2020, Yaakoubi2021,tahir2021}. Graph-native machine-learning predictors have been employed, such as convolutional neural networks~\citep{Yaakoubi2020,tahir2021}, deep structured predictors~\citep{Yaakoubi2021}, pointer networks~\citep{Bello2017}, graph neural networks~\citep{morabit2021}, and graph convolutional networks~\citep{gasse2019}. In short, this line of work has shown that graph-based predictors proved to be an effective framework for learning graph-structured data and capturing dependencies between nodes, which provided information for OR solvers to optimize problems better and faster~\citep{Cappart2021}. Another independent stream of research has focused on graph-based reasoning for asset valuation in financial systems, modeling the interactions between financial assets. However, to the best of our knowledge, no previous research has examined the use of graph-based reasoning and representation learning to directly solve large-scale stochastic industrial asset valuation problems.

\vspace{-0.3cm}
\paragraph{Contributions:} This paper proposes to fill the gap in the current literature by proposing a solution methodology that combines AI and OR for SSOMC, as well as introducing an end-to-end graph-based mineral asset valuation tool under geological and commodity price uncertainty. A graph-based reasoning model is first constructed to capture the relationships between grade and material representations in the ground, ore body block attributes, and mining complex components, in order to learn a block-sampling representation portraying the effect on production in downstream optimization and finished product prices. This model takes as input a graphical representation of the ore body models and interacts with a hyper-heuristic solver in a reinforcement learning (RL) setting to actively learn to predict the outcome of perturbing a certain ore body block on the overall schedule, thus on the NPV. The ore body models are represented by a graph-structured simulated dataset, and the commodity price uncertainty is represented by a time-series dataset. The mining complex is modeled as a graphical structure that incorporates all transformations and interactions in the mineral supply/value chain. The rest of the paper is structured as follows: Section~\ref{prob} presents an overview of the problem formulation and the uncertainty modeling approach, while section~\ref{meth} presents the proposed methodology. Section~\ref{comp_exp} presents computational experiments on two mining complexes producing respectively copper and gold, followed by conclusions in Section~\ref{conc}.

\vspace{-2mm}
\section{Problem Statement and Formulation}\label{prob}
\vspace{-2mm}

\begin{figure}[t!]
\centering
\includegraphics[width=0.8\linewidth]{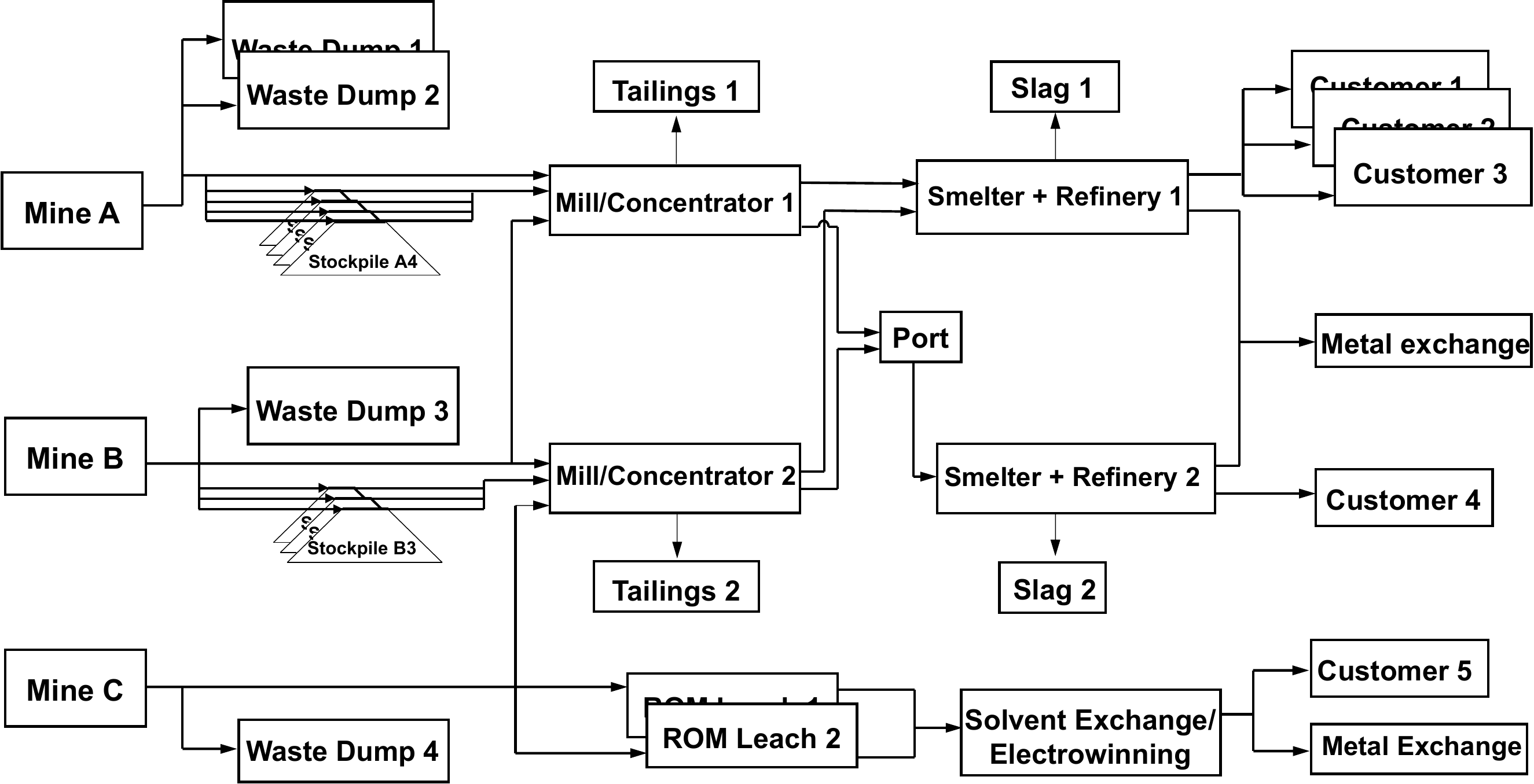}
\vspace{-2mm}
\caption{An illustration of a mining complex~\citep{goodfellowThesis,goodfellow2016}.}
\vspace{-5mm}
\label{fig:3}
\end{figure}

A mining complex~\citep{goodfellowThesis, goodfellow2016} typically contains a set of mines $m \in \mathbb{M}$, a set of stockpiles $\mathbb{S}$, and a set of processors $p \in \mathbb{P}$. Each mine is discretized into ore body blocks $b \in \mathbb{B}_m$ characterized by their material type and attribute values $\beta_{p,b,s}$, where $p \in \mathbb{P}$ refers to primary attributes and $s \in \mathcal{S}$ refers to a scenario in the set of scenarios $\mathcal{S}$ used to represent the combined uncertainty in grade, related geochemical properties, and commodity prices. Material flow is modeled using a graph structure with nodes $\mathcal{N}$ (sources/destinations) connected by arcs representing allowable incoming-outgoing pairs $(\mathcal{I}(i) \subset \mathcal{N}, \mathcal{O}(i) \subset \mathbb{S} \cup \mathcal{P} )$. As shown in figure~\ref{fig:3}, the material extracted from mine flows through different processes, propagating the uncertainty to all operations in the mining complex, which culminates in financial risk. As such, SSOMC optimizes production scheduling decisions, destination policies, processing stream decisions, and transportation alternatives in a single mathematical model, integrating various nonlinear material transformations and interactions. The framework is particularly flexible, allowing nonlinear calculation of attributes (blending, grade recovery function, geometallurgical responses) at any stage of the mining complex, thus hindering the use of any exact solution methods. The problem can be modeled using a two-stage stochastic integer program. As in equation~\eqref{eq:3}, the objective function maximizes the expected NPV and minimizes the expected recourse costs incurred in case of violation of the stochastic (scenario-dependent) constraints. The parameter $p_{h, i, t}$ denotes the time-discounted price (or cost) per unit of attribute, while $v_{h, t, s}$ is used to account for the quantity/value of non-additive hereditary attributes $h$ (e.g., metal grade) as material moves through the value chain. Scenario-independent constraints include reserve constraints, slope/precedence constraints, and mining capacity constraints. Scenario-dependent constraints typically include processing capacity and stockpiling constraints.

\vspace{-4mm}
\begin{equation}\label{eq:3}
\max \frac{1}{|\mathcal{S}|} \underbrace{\sum_{s \in \mathcal{S}} \sum_{t \in \mathbb{T}} \sum_{h \in \mathbb{H}} p_{h, i, t} \cdot v_{h, i, t, s}}_{\text {Discounted costs and revenues }}-\frac{1}{|\mathcal{S}|} \underbrace{\sum_{s \in \mathcal{S}} \sum_{t \in \mathbb{T}} \sum_{h \in \mathbb{H}} c_{h, i, t}^{+} \cdot d_{h, i, t, s}^{+}+c_{h, i, t}^{-} \cdot d_{h, i, t, s}^{-}}_{\text {Penalties for deviations from targets }}.
\end{equation}
\vspace{-4mm}

In this work, the SSOMC~\citep{goodfellowThesis,goodfellow2016} is solved under supply and demand uncertainties. The supply is incorporated as equiprobable scenarios of the attributes of interest (such as grades, material types, etc.).  To model the market uncertainty, metal price simulations are generated using stochastic reduced-form models. Copper can be simulated using a \textbf{mean reverting process with Poisson jump diffusion} as in equation~\ref{eq:1}, where $S_t$ is the price of the metal at time $t$, $\alpha$ is the commodity’s mean-reverting speed, $W$ is a Wiener process, and $\beta$ is the jump size of Poisson process $P$. Gold can be simulated using a \textbf{trending geometric Brownian motion model with Poisson jump diffusion}~\citep{schwartz1997} as in equation~\ref{eq:2}, where $\eta$ is the average annual price drift and $\sigma$ is the average annual volatility.

\vspace{-4mm}
\begin{equation}
S_t = S_{t-1} + \alpha \cdot (\bar{S} - S_{t-1}) + W \cdot S_{t-1} + \beta \cdot P \cdot S_{t-1}
\label{eq:1}
\end{equation}
\vspace{-4mm}
\begin{equation}
S_t = S_{t-1} \cdot exp(\eta \cdot t - \sigma^2 \cdot \frac{t}{2} + W + \beta \cdot P)
\label{eq:2}
\end{equation}
\vspace{-7mm}

\section{Proposed Solution Methodology}\label{meth}

\begin{figure}[!htb]
\centering

\centering
\scalebox{0.48}{
\begin{tikzpicture}   

  \node(poli)[twoblock]{\nodepart[text height=1cm]{one} Guiding policy
               \nodepart{two}
                \begin{tikzpicture}
                \node(poli6)[crect] {Generate a heuristic\\sampling distribution\\$S_{\theta}(h)$};
                  \end{tikzpicture}
            
\begin{tikzpicture} \node(poli5)[crect,right= 1mm of poli6] {Update agent's\\weights $\theta$};   
   \end{tikzpicture}};
   
   \node (c5)[circmarker] at (poli.north west) {5};

   \node(past)[twoblock, right=5mm of poli]{\nodepart[text height=1cm]{one} \quad Past Experience
   \nodepart{two} Store experience \\in replay buffer};
   \node (c4)[circmarker] at (past.north west) {4};

 \draw[->,very thick] (past.west) -- (poli.east); 
 
 \node(Reinempty)[below=7mm of poli]{};
 
  \node[draw,fit=(Reinempty) (poli) (past)](Rein) {};
 
  \node(Reintit)[above=0cm of Rein.south, align=center,draw,fill=black!20,inner sep=3mm]{Neural Guiding };
 %
 %
  \node(samp)[twoblock, below=28.5mm of poli]{\nodepart[text height=1cm]{one} \rule{0mm}{4mm} Heuristic sampling function
               \nodepart{two}
               Update the sampling function};
 \node (c7)[circmarker] at (samp.north west) {6};             
 

\draw[->,very thick](poli.west) |-  ++(-5mm,0) |- (samp.text west);

  \node(stoch)[twoblock, below= 5mm of samp]{\nodepart[text height=1cm]{one} Stochastic Heuristic: Selection Mechanism
               \nodepart{two}
                
                \begin{tikzpicture}
                \node(stoch1)[crect] {Select a\\heuristic $h_i$\\ };
                  \node (c1)[circmarker] at (stoch1.north west) {1};
                  \end{tikzpicture}
                  
\begin{tikzpicture} \node(stoch2)[crect,right= 1mm of stoch1] {Apply the\\selected heuristic\\ $h_i(x)$};   
\node (c2)[circmarker] at (stoch2.north east) {2};            
   \end{tikzpicture}          };

  \node(simu)[twoblock, right=12mm of stoch]{\nodepart{one} \quad Simulated Annealing Acceptance Criterion
   \nodepart[align=left]{two} \framebox{$x' \gets h_i(x)$} accept the selected heuristic\\[1.2mm]
    \hspace{29mm} OR\\[1.2mm]
     \framebox{$x' \gets x$} \quad reject (restore)};
  \node (c3)[circmarker] at (simu.north west) {3};

   \node (c4)[circmarker] at (past.north west) {4};
    
\draw[->,very thick] (stoch.east) -- (simu.west);

\node(stop)[decision,left= 18 mm of stoch]{Stop?};

\node(yesnode1)[above=10mm of stop.west]{};
\node(yesnode2)[left=-4mm of yesnode1.west]{Yes};

\node(weststop)[left= 9mm of stop]{};

\node(St)[circstate,left= 7 mm of stoch, label=below: {$x$}] {};

\draw[->,dotted, very thick](stop.east)-- node[above]{No}(St.west);
\draw[->,very thick](St.east)--(stoch.west);

\draw[->,very thick] (samp.two west) |-  ++(-15mm,0mm) -|(stop.north);

\draw[->,very thick, dotted](stop.north west)-|++(-2mm,0mm)|- ++(0mm,25mm) node[above](OBs){\begin{tabular}{c}Output\\ Best \\solution
\end{tabular}};


\node(coll)[draw,very thick, below=21mm of simu.west]{\begin{tabular}{l}Collect problem independent information for domain barrier (e.g., the number of \\heuristics, the changes in evaluation function, a new solution or not, the distance \\between two solutions, etc.)
\end{tabular}};

\node(darr1)[double arrow, inner sep=1mm,  shape border rotate=90, double arrow head extend=1mm, minimum height=6.7mm,draw, below=0mm of stoch.south]{};

\node(darr2)[double arrow, inner sep=1mm,  shape border rotate=90, double arrow head extend=1mm, minimum height=7mm,draw, below=0mm of simu.south]{};

\node(Simulated)[below=7mm of coll]{};

\node[draw,fit= (samp)  (weststop)(simu) (coll) (Simulated) (OBs)](Simu)  {};
 
 \node(Simutit)[above=0cm of Simu.south, align=center,draw, fill=black!20!white,inner sep=3mm]{ Simulated Annealing };

\node(St1)[circstate,right= 3 mm of Simu, label=below: {$x'$}] {};
\draw[->,very thick, dotted](simu.east) |- (St1);
\node(Doma)[below=3mm of Simu, draw, fill=black!20!white, text width=20cm, text height=2.5mm, align=center, inner sep=3mm]{Domain Barrier};

\node(H1)[circH, below=70mm of St]{$H_1$};
\node(Hn)[circH, right=12mm of H1]{$H_n$};
\node(H2)[circH, below=2mm of H1, xshift=2mm]{$H_2$};
\node(itd)[right=7mm of H2]{\ldots};

\begin{scope}
 \tikzset{every fit/.style={text width=2cm, inner sep=-1mm}}

\node[ellipse,minimum height=2mm, minimum width=51mm,draw,thick, fit=(H1) (H2) (itd) (Hn)] (HH){};
\end{scope}

\node(PEI)[draw,right=40mm of HH,align=left, thick, text width=50mm]{\\
\hspace{4mm}-- Problem representation\\
\hspace{4mm}-- Evaluation function\\
\hspace{4mm}-- Initial solution ($S_0$)\\
};

\node(S0)[circstate,left= 15 mm of HH, label=below: {$S_0$}] {};

\draw[->,thick] (S0) |- (stop);

\node(Problem)[below=5mm of HH]{};

\node[draw, fit=   (HH) (PEI) (Problem) ](Prob)  {};

 \node(Probtit)[above=0cm of Prob.south, align=center,draw, fill=black!20!white, inner sep=3mm]{Problem Domain};
 
\node(relative)[right= 35 mm of Simu] {};
\node(branch)[twoblock, right=11cm of samp, yshift=-3cm,inner sep=3mm]
{\nodepart[text height=1cm]{one} Branching Policy 
               \nodepart{two}
                \\[15pt]
                \begin{tikzpicture}
                \node(gene1)[crect] {Generate a block-\\sampling distribution\\$S_{\theta'}(b)$};
                
                 \node(gene2)[crect,below= 35mm of gene1] {Update agent's\\weights $\theta'$};
                 \draw[->,very thick] (gene2.north) to (gene1.south);
                  \end{tikzpicture}\\[5pt]};
\node (c5)[circmarker] at (branch.north west) {4}; 

\node(coll2)[draw,very thick, right=12 mm of Prob, yshift=10mm]{\begin{tabular}{l}Collect problem-independent\\ information (e.g., orebody \\representation, block\\attributes, etc.)
\end{tabular}};

\node(neural)[below=14mm of coll2]{};
\node[draw, fit= (branch) (coll2) (neural),text width=4.8cm,inner sep=3mm ](Neu)  {};

\node(Neutit)[above=0cm of Neu.south, align=center,draw, fill=black!20!white, inner sep=3mm]{Neural Branching};
\draw[->,very thick]([yshift=18mm]branch.west) |- ++ (-12mm,0) |- (past.east); 
\draw[->,very thick](St1) -- (St1-|branch.west);
\node(darr3)[double arrow, inner sep=1mm, double arrow head extend=1mm, minimum height=7mm,draw, left=4mm of coll2.west]{};
\node(darr4)[double arrow, inner sep=1mm, double arrow head extend=1mm, minimum height=7mm,draw, right=16mm of coll.east,yshift=2mm]{};
 \end{tikzpicture}
} 

   
\caption{Illustration of the proposed solution methodology.}
\vspace{-5mm}
\label{Flowchart}
\end{figure}
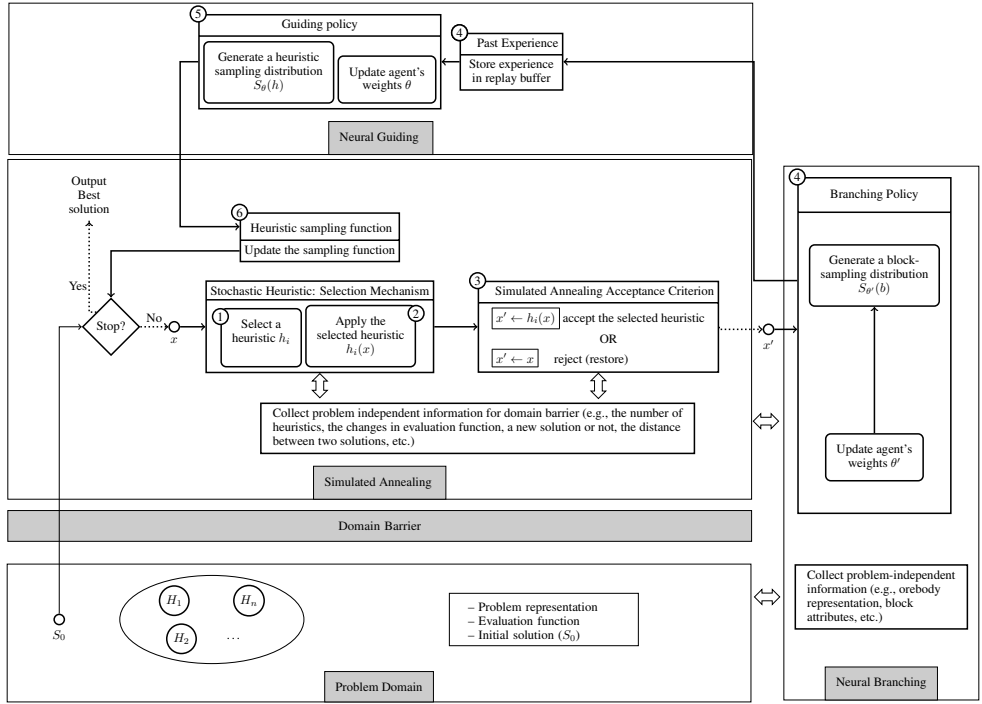

The proposed solution methodology consists of four components, as in figure~\ref{Flowchart}: the problem domain containing the ore body representation, the hyper-heuristic managing the value/supply chain optimization and dynamics via a pool of heuristics, a neural guiding policy exploring a heuristic selection tree, and a neural branching policy learning a latent mineral representation of the blocks that maps the simulated ore body blocks to their effect on the overall mineral value assessment.

\vspace{-1mm}
\subsection{Problem domain}
\vspace{-1mm}
The problem domain includes the problem representation, the evaluation function, the initialization procedure, and a set of low-level heuristics. These heuristics work to make small changes (perturbations) to the current solution and search in the following three neighborhoods:
\begin{itemize}
    \setlength\itemsep{0mm}
    \item Block extraction sequence perturbations: a block's extraction period is changed, according to the heuristic's rules, and any potential slope constraints that were violated are fixed.
    \item Destination policy perturbations: the destination of a randomly selected group of blocks is changed. Blocks with similar attributes are sent to the same destination.
    \item Processing stream perturbations: A random normal number is used to change a randomly selected process stream variable, thus perturbing the supply/value chain variables (dynamics).
\end{itemize}

\subsection{Hyper-heuristic}
The hyper-heuristic adapted from~\citep{goodfellowThesis,Yaakoubi2022_2,yaakoubi2022} has access to a pool of heuristics and iterates between perturbing the current solution $x$ and assessing whether to accept or reject the perturbed solution. Once a heuristic $h_i$ is selected, $x$ is modified into a new solution $x'$. The selection of the heuristic is based on heuristic sampling distribution $S_{\theta}(.)$, where $\theta$ are the weights parameterizing a guiding policy-based RL agent. Then, a simulated annealing acceptance criterion is used with the probability defined as in equation~\eqref{Eq:G6}, where $\Delta f(h_i)$ is the difference in the objective function value between $x$ and $x'$, and $\Temp$ is the simulated annealing temperature.

\vspace{-3mm}
\begin{equation}
p(h_i(x)) =
\begin{cases}
$1$ & \text{if $\Delta f(h_i) > 0$ }\\
\exp \left(-\dfrac{\Delta f(h_i)}{\Temp}\right) & \text{otherwise}
\end{cases}
\label{Eq:G6}
\end{equation}
\vspace{-6mm}

\subsection{Neural Guiding}

In order to guide the search towards more promising regions of the heuristic space, an approach similar to the learn-to-perturb (L2P) hyper-heuristic~\citep{yaakoubi2022,Yaakoubi2022_2} is adapted. In short, a heuristic selection strategy can be viewed as a scoring function $f$ that outputs a score for each branching candidate. As such, $f$ can be modeled as a parametric function and optimized by an RL algorithm. 

\textbf{Problem Definition:} Given a list of heuristics $h_i$ ($i = 1$, \ldots, $n$), an agent parameterized by its weights $\theta$, and recent heuristic performance for the last few iterations, the neural guiding agent aims to propose a sampling distribution $S_\theta(.)$, so that the sequence of heuristics sampled improves the objective function value (as in equation~\eqref{eq:3}) in the minimum amount of time required. This heuristic selection mechanism defines a \textit{tree structure} where each node is identified by the heuristics performed so far and the solution at that given node, and the next node is partially identified by the score function computed at the node.
Note that since the RL agent searches the heuristic space rather than the solution space, the environment is considered a partially observable Markov decision process (MDP).

\textbf{State and action:} At time step $t$, the state $s_t$ is defined as the performance of the heuristics for the last few iterations (nodes). Thus, the input is a concatenation of all performance metrics (e.g., improvement in the objective function, execution time, etc.) for all heuristics for the last few iterations. The agent outputs a vector in $[0,1]^n$, conveying the predicted future performance of each heuristic.
In contrast to previous work~\citep{yaakoubi2022,Yaakoubi2022_2} where the output normalization is performed directly on the output actions, a Gaussian noise $N(0,\sigma)$ is first added to encourage exploration, then a $L1$ vector normalization is performed to compute the final score function. The agent is, thus, provided the possibility to change its vector norm depending on how confident it is on the predicted score.

\textbf{Reward:} The reward at time step $t$ is defined in equation~\eqref{eq:reward}, where $h_i$ is the heuristic selected using the score function $S_\theta(.)$, $T(h_i)$ the time required to use the heuristic $h_i$, and $\Delta f(h_i)$ the optimization gain. In contrast to previous work~\citep{Yaakoubi2021,Yaakoubi2022_2}, the execution time is included in the reward, as no voting mechanism is used to balance the agent's output with a predefined score function.

\vspace{-4mm}
\begin{equation}
r_t = \frac{\Delta f(h_i)}{T(h_i)} \mathbbm{1}_{\Delta f(h_i) \geq 0} + \Delta f(h_i) \cdot T(h_i) \cdot \mathbbm{1}_{\Delta f(h_i) < 0}
\label{eq:reward}
\end{equation}
\vspace{-2mm}

\textbf{Model architecture and optimization:} Since the model is a partially observable MDP, sample efficiency can be a major issue when deploying the solver, especially since internal training instances may not be available at test time (due to privacy concerns). Thus, the policy-gradient algorithm in~\citep{Yaakoubi2022_2, yaakoubi2022} is replaced by a variant of the Q-learning algorithm, Rainbow~\cite{rainbow2}, a model-free value-based RL algorithm. Rainbow is shown to speed up learning compared to latest policy-gradient algorithms used in previous work~\citep{Yaakoubi2022_2, yaakoubi2022}, avoiding the need to warm-start or pre-train.

\subsection{Neural branching}

To the best of our knowledge, in the existing literature on mine planning and optimization and mineral value assessment, all approximate solution methodologies sample blocks to be perturbed in a random fashion. Although these methods provide a general approach that requires no fine-tuning, simulated ore body blocks may not have the same effect on the objective function value (equation~\eqref{eq:3}). In fact, we argue that some ore body blocks have a greater effect on production scheduling and downstream optimization, and thus on the mineral value assessment. This effect may be due to block location, metal content, etc. More importantly, the effect of a block may also depend on its neighboring blocks. In other words, the contribution of a simulated ore body block to NPV is not only local but also holistic. This paper proposes the first mineral ore body representation learning that is integrated in a planning and optimization framework. The proposed representation is used to map block attributes to mineral value assessment while harnessing the graph-based structure of the learning problem.

\textbf{Problem Definition:} Given a mineral deposit represented as a 3D graph, the neural branching agent aims to generate a block-sampling function $S_{\theta'}(b)$ that would be used by low-level heuristics to perturb the extraction sequence, where $\theta'$ are the weights parameterizing the agent. This sampling function should allocate a higher probability to blocks that would improve the objective function value in a minimum amount of time, and a low (or zero) probability to blocks that would deteriorate the objective function value or take a long time when perturbed by the heuristic.

\textbf{Graph:} The ore body model can be considered a graph $G = (V,E)$ where $V$ is the set of nodes, and $E$ is the set of edges. Each node $v \in V$ corresponds to an ore body block and is characterized by the attributes of a block (e.g., weight, ore tonnage, extraction period, etc.). Two nodes are connected if their corresponding ore body blocks are adjacent. Next, three different methods are proposed to actively learn a mapping between a graph-structured input (ore body model) and the objective function value (equation~\eqref{eq:3}), trained with a model-free policy gradient algorithm, PPO ~\citep{ppo}.

\begin{figure}[t]
\centering
\includegraphics[width=.48\linewidth]{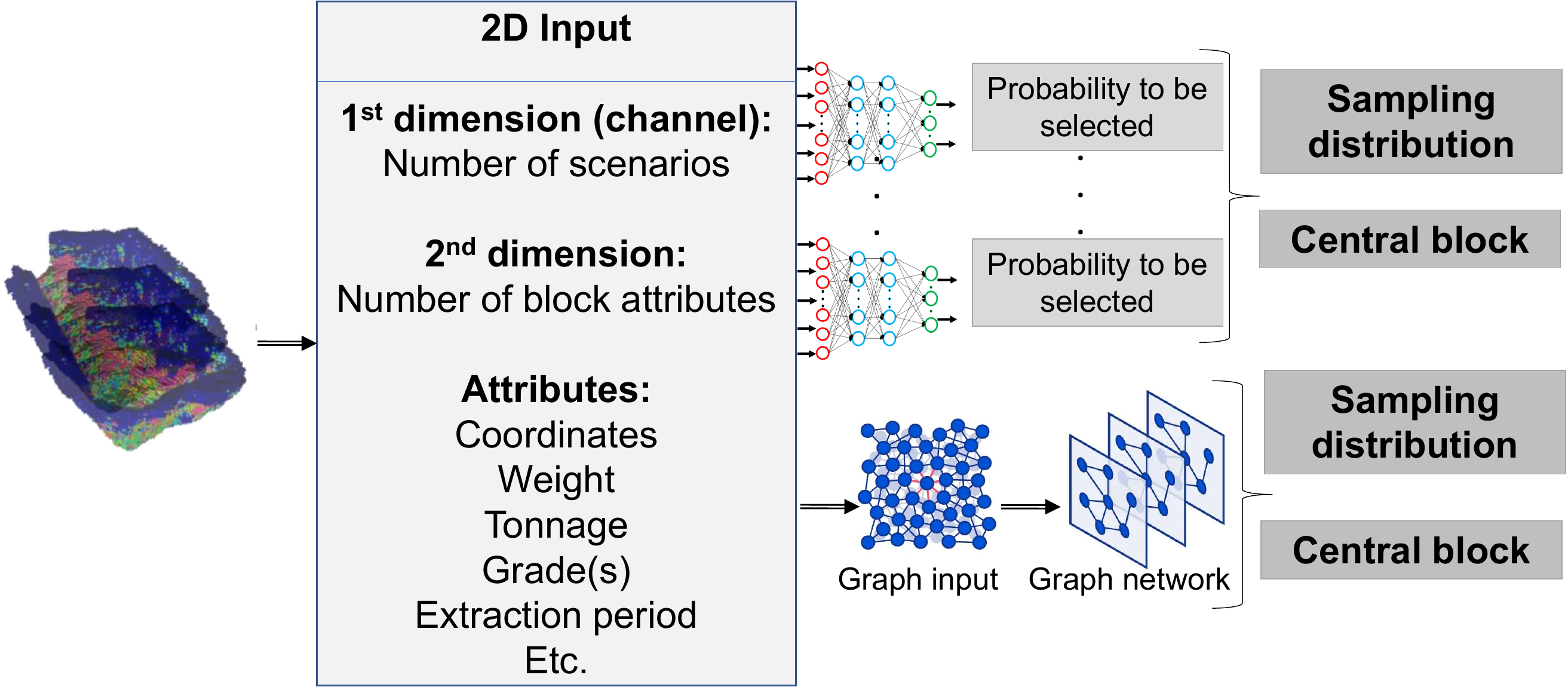}
\hspace{3mm}
\includegraphics[width=.48\linewidth]{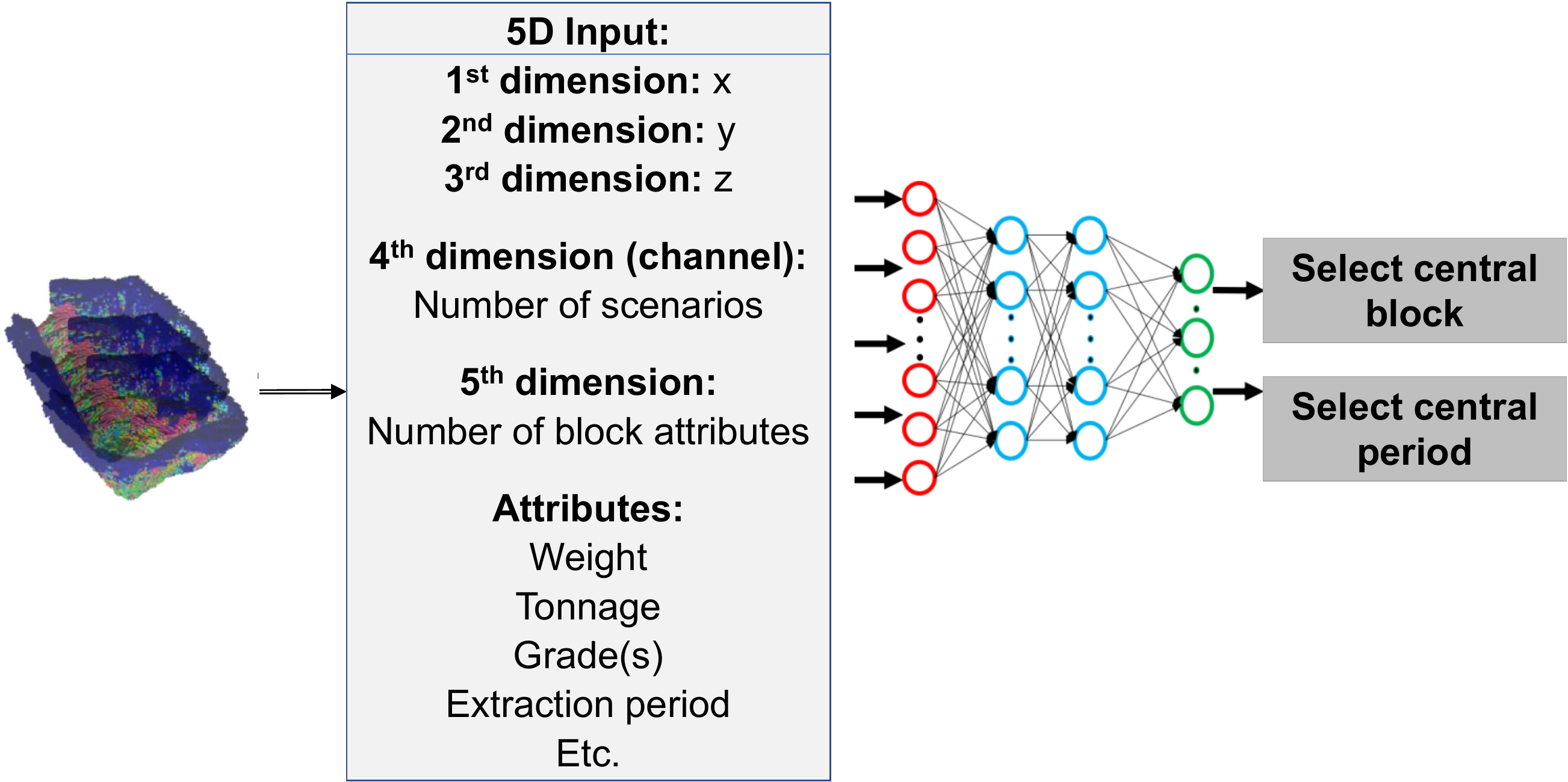}
\caption{Input and network architecture for the neural branching policy using the block-based (left) and radius-based neural branching (right), referred to as NN-NB and RB-NB, respectively.}
\label{fig:l2b2}
\vspace{-5mm}
\end{figure}

The first method, shown in figure~\ref{fig:l2b2}, is the block-based neural branching (\textbf{NN-NB}) which constructs an input per block, then passes the inputs in parallel to the same neural network producing a probability for each block to be selected. These probabilities are normalized to make the final sampling distribution $S_{\theta'}(b)$. This weight sharing technique is used to reduce the number of parameters to learn, which has been shown to accelerate the learning.
The second method, radius-based neural branching (\textbf{CNN-NB}), uses a convolutional neural network (CNN) as in figure~\ref{fig:l2b2}, to output the coordinates of a central block and constructs a block-sampling function that has the highest probability density function at the identified central block.
In this context, CNNs have proven useful in a graph-structured input to discover local patterns by using a convolutional layer to extract local neighborhood features from each node in the graph, and then using the pooling layer to reduce the dimensionality of the data.
The third method (\textbf{GNN-NB}) uses graph convolutional neural networks (GCNNs) to generate a block-based sampling probability. GCNNs are designed to iteratively update the representation of a node (also called state) by aggregating information from neighboring nodes. As the iterations progress, the nodes collect information from far neighbor nodes. In this work, the graph mentioned above is directly fed to the GCNN, and a probability is generated per node (block).

\section{Experiments}\label{comp_exp}

\begin{table}[b]
\caption{Parameters used to model metal price uncertainty algorithms.}\label{table1}
\vspace{-2mm}
\centering
\resizebox{0.8\linewidth}{!}{
\begin{tabular}{ll}
\toprule
Parameter & Value and Description\\
\midrule
Initial value, $S_0$ and reversion level , $\bar{S}$& US\$2.78/lb \\
Annual volatility, $\sigma$ & 9\%, average annual volatility over $25$ years\\
Mean reverting speed $\alpha$ & $0.5$ \\
Average jump frequency $\mu_p$ & $1$ per year, $25$-yr average number of Cu price shocks \\
Average jump size, $\beta$ & 3\%\\
\midrule
Initial value, $S_0$ & US\$1548.6/oz, price assumption used by operation \\
Annual volatility, $\sigma$ & $5\%$, average annual volatility over $25$ years \\
Annual drift, $\eta$ & $0.1\%$, $5$-year moving average drift over $25$ years \\
Average jump frequency, $\mu_p$ & 1 per year \\
Average jump size, $\beta$ & $0.5\%$ \\
\bottomrule
\end{tabular}
}
\end{table}

The parameters used to model the price uncertainties are shown in table~\ref{table1}. The proposed framework is tested on two large-scale industrial mining complexes: a copper and a gold deposit, and compared to a state-of-the-art solver for SSOMC~\citep{goodfellowThesis, goodfellow2016}. The baseline uses simulated annealing with a predefined heuristic selection function, and has been adapted to use the same set of heuristics as in this work. The main differences between the baseline and the proposed approach are the heuristic operation (informed by neural branching in this work) and the heuristic selection mechanism (through neural guiding). In what follows, NN-NB, CNN-NB, and GNN-NB refer to the proposed solution using neural guiding and deploying block-based, radius-based, and GNN-based neural branching, respectively.

\begin{figure}[ht]
\centering
\subfigure[Copper Mine]{
\includegraphics[width=.36\linewidth]{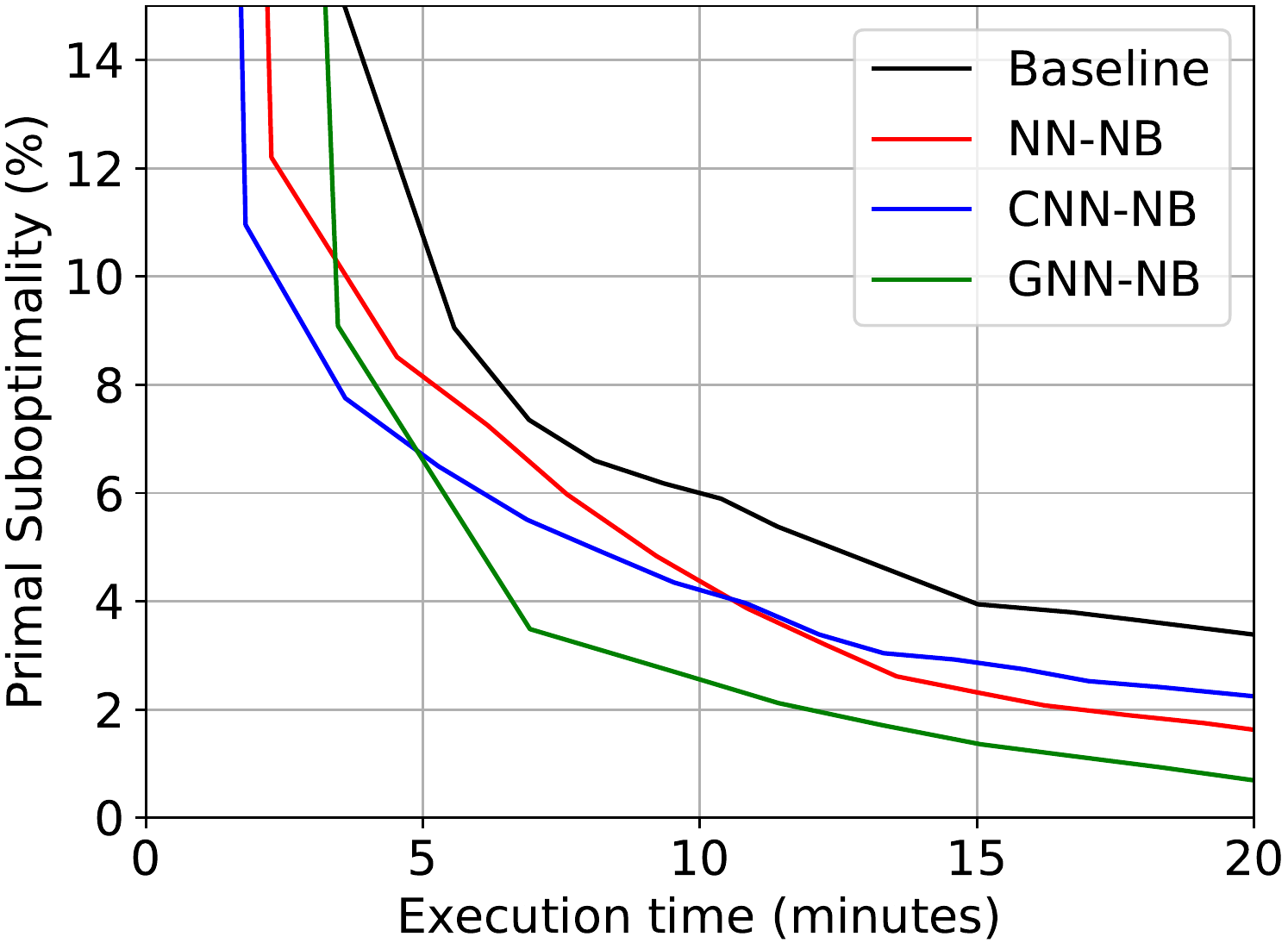}
\hspace{2mm}
\includegraphics[width=.36\linewidth]{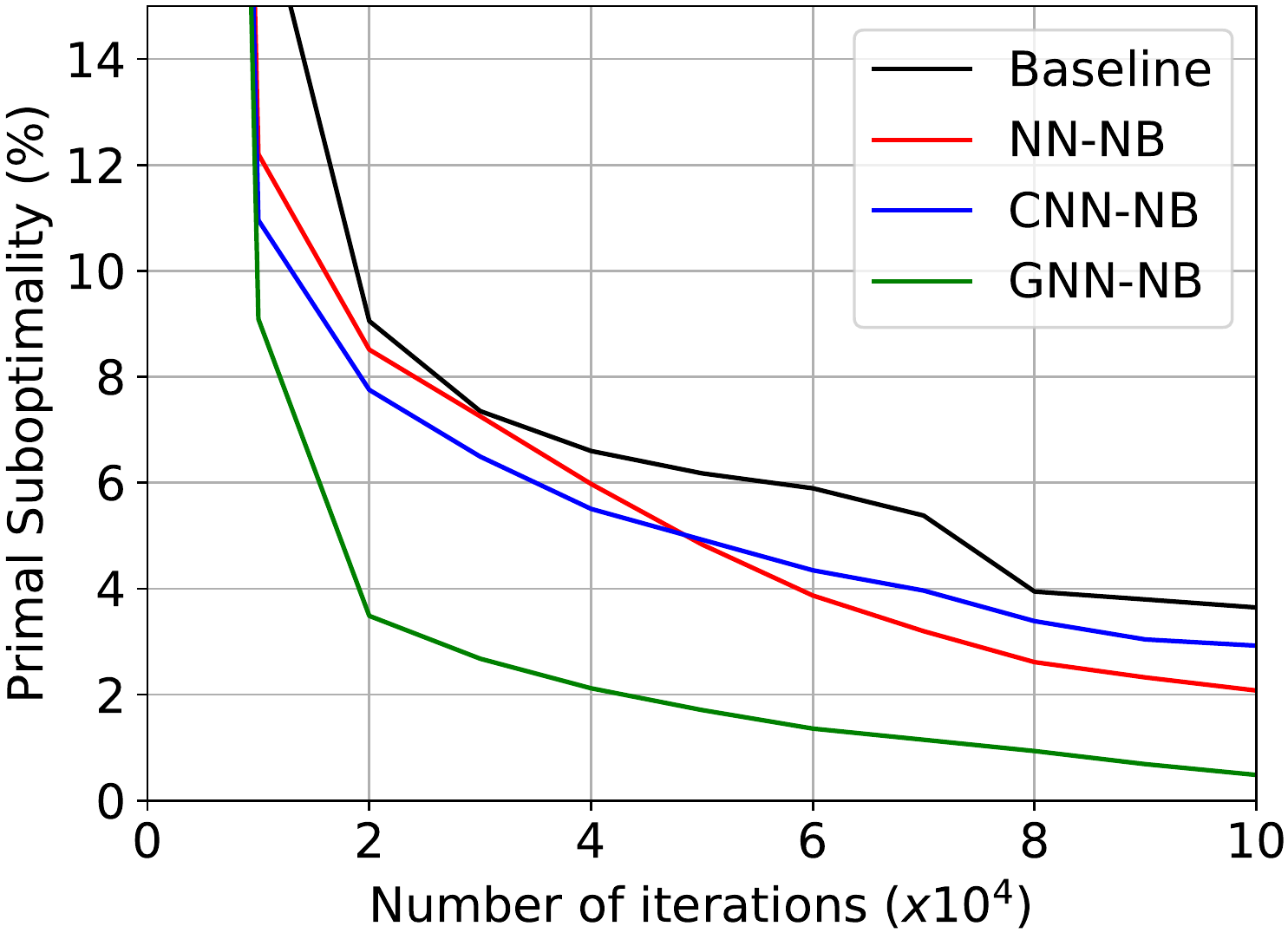}}
\vspace{-3mm}
\subfigure[Gold Mine]{
\includegraphics[width=.36\linewidth]{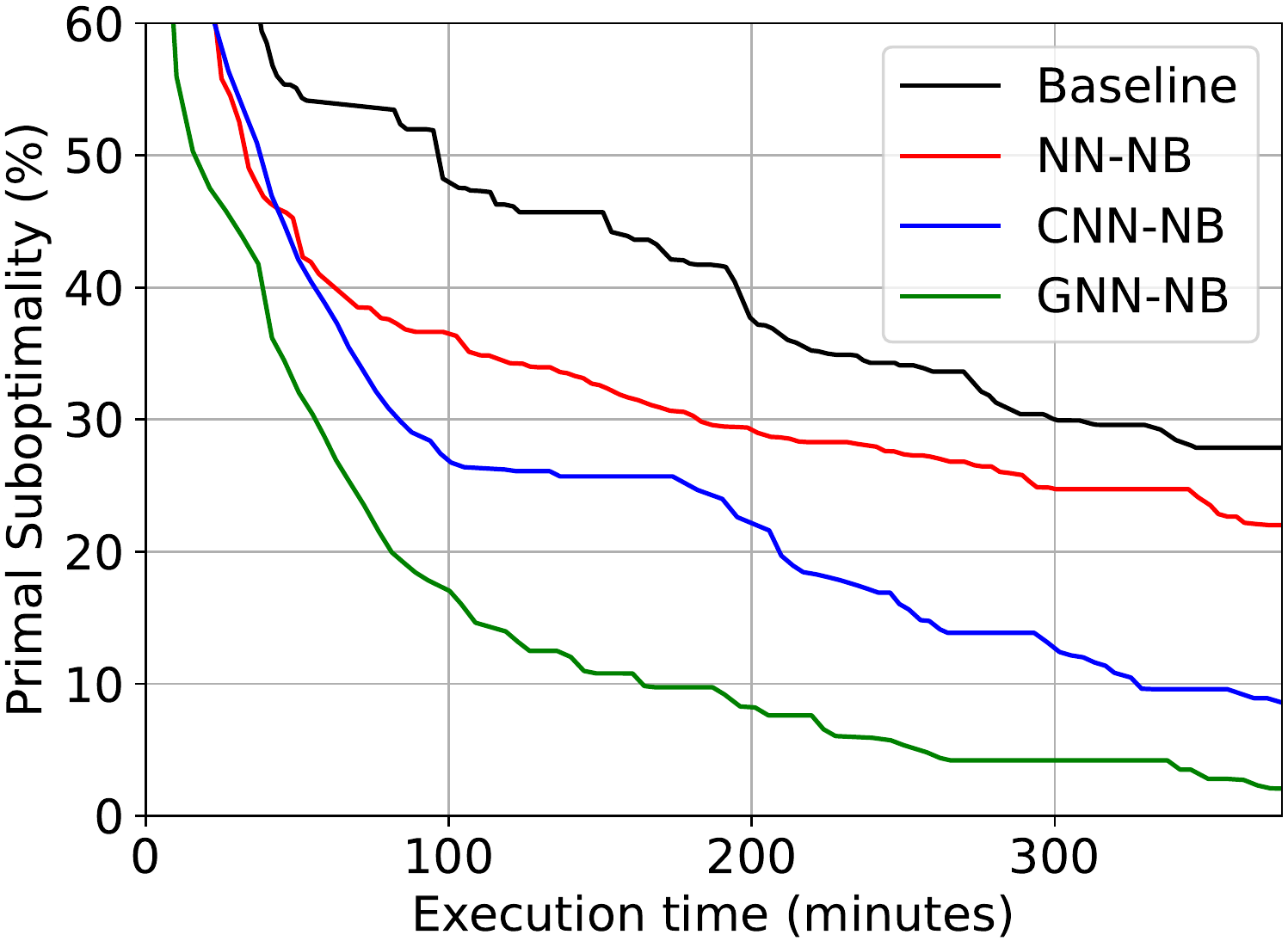}
\hspace{2mm}
\includegraphics[width=.36\linewidth]{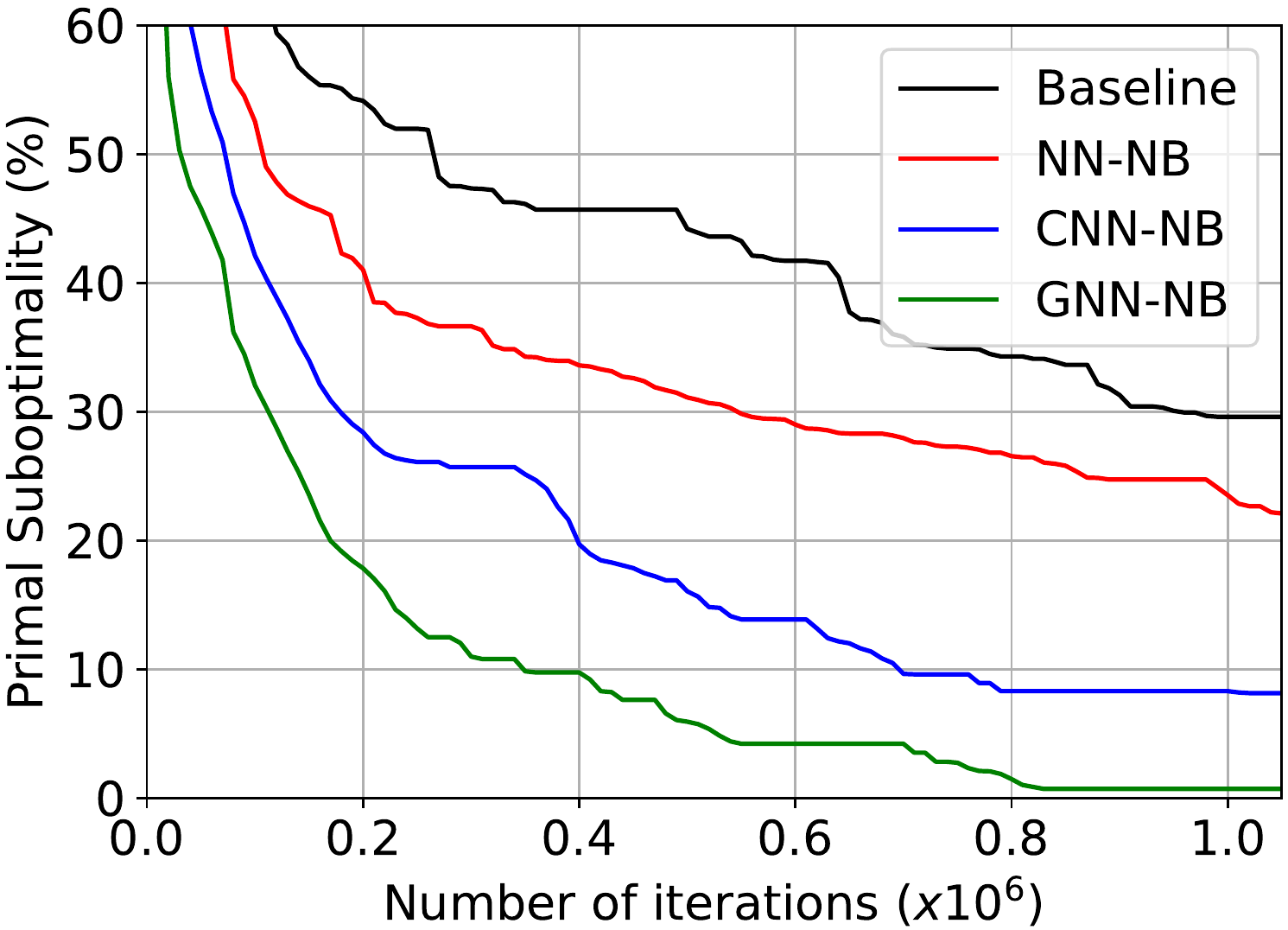}}
\caption{
Primal sub-optimality progress for (a) the copper mine, and (b) the gold mine.
}
\label{fig:primalSub}
\vspace{-5mm}
\end{figure}

The first case study is a copper mine containing $200$k ore body blocks and the time horizon is $8$ years. The model incorporates $20$ simulated ore body realizations and $1000$ realizations for the price uncertainty. The number of variables is approximately $2 \cdot 10^6$. The second case study is a gold mine containing $1.2$ million blocks and the time horizon is $15$ years. The model incorporates $15$ simulated ore body realizations and $1000$ realizations for the price uncertainty. The number of variables is approximately $18.5 \cdot 10^6$. The discount rate for the cash flow is $10\%$.

For comparison, the primal suboptimality progress is reported in figure~\ref{fig:primalSub}, which is computed using the progress of the objective function value as in~\eqref{eq:3}.
The progress is shown as a function of the number of iterations and the execution time (in minutes). While all the proposed variants outperform the baseline~\citep{goodfellow2016}, GNN-NB outperforms the other variants. For instance, in the copper mine, GNN-NB reduces the primal suboptimality by up to $20$x, the execution time by up to $70\%$, and the number of iterations by up to $80\%$. Far better significant performances are found for the gold copper deposit, where GNN-NB reduces the primal suboptimality by up to $2000$x, and reduces the execution time and number of iterations by three orders of magnitude.

\begin{figure}[htb]
\centering
\includegraphics[width=.36\linewidth]{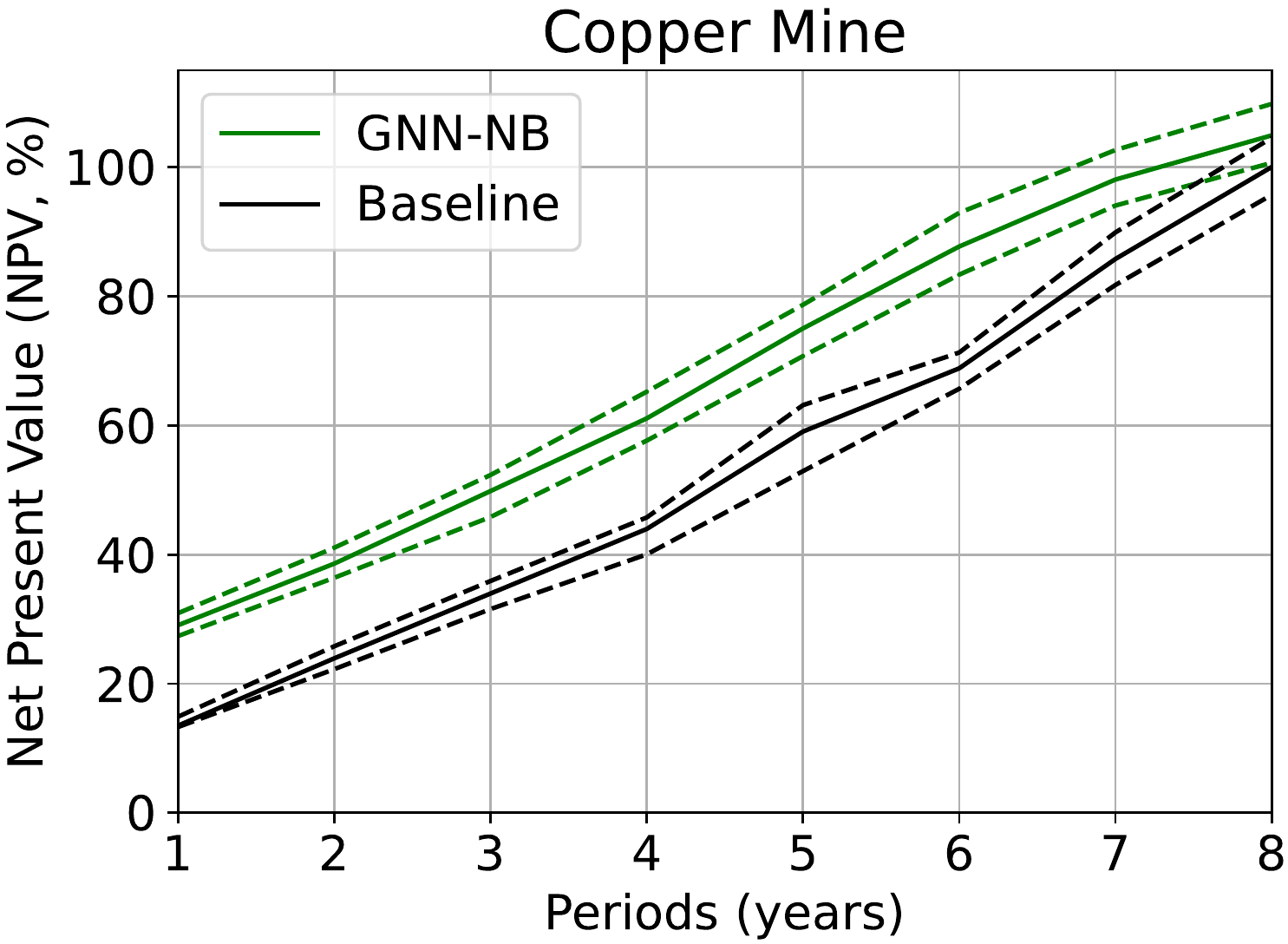}
\hspace{3mm}
\includegraphics[width=.36\linewidth]{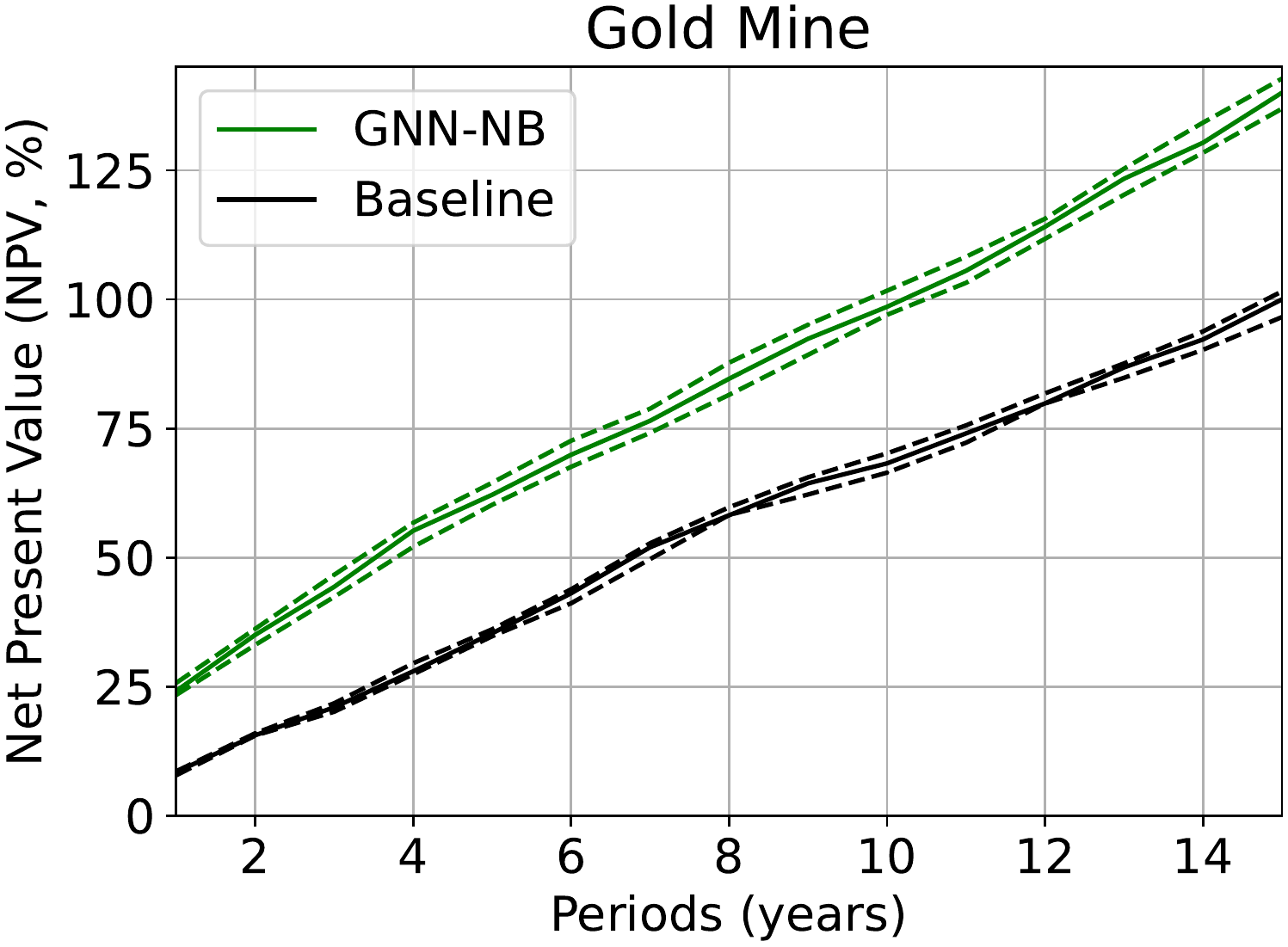}
\vspace{-2mm}
\caption{Mineral asset valuation for the copper and gold mines from optimized schedules obtained with the baseline and the GNN-NB solution.}
\vspace{-2mm}
\label{fig:npv}
\end{figure}

Next, the optimized schedules from the baseline and GNN-NB obtained using the same run time (time-based early stopping) are used to compare the mineral asset valuations. Figure~\ref{fig:npv} reports the NPV for the copper and gold deposits as confidence intervals for the estimated P10, P50 and P90 quantiles. For the copper deposit, GNN-NB provides a mineral asset value that is $5\%$ higher than the baseline, which is equivalent to a multi-million dollar gain. For the gold deposit, the improvement is more significant, with the mineral asset value found being $40\%$ higher than the baseline, equivalent to a gain of several hundred million dollars. The difference in performance is due to two reasons: (1) the size of the mining complex, so the more complex the problem, the longer it takes to solve, and the more effective a solution methodology will be on the final schedule and valuation, (2) the effect of the learned ore body representation (block-based sampling function) on the downstream optimization. If the ore body is sparse (high-grade ore blocks are scattered), more blocks must be sampled to find a good sampling strategy. If the ore body is dense (many portions will be unmined, mined to access the ore, or used for blending), then the distinction between blocks that need to be sampled and those that will have no effect on the optimization progress becomes much easier to learn.
\vspace{-1mm}
\vspace{-1mm}
\section{Conclusions}\label{conc}
This work uses graph-based reasoning, modeling, and learning to propose an end-to-end graph-based mineral asset valuation tool under geological and commodity price uncertainty. The proposed methodology uses a neural guiding policy to explore a heuristic (perturbation) selection tree and a neural branching policy to learn a block sampling representation of the mineral body. Results show a significant reduction in primal suboptimality and execution time by up to three orders of magnitude, as well as a significant increase in mineral asset valuations for two large-scale industrial mining complexes. The proposed approach can be used for any graph-based combinatorial optimization problem, including supply/value chains, airline and railway scheduling, and vehicle routing problems.

\bibliographystyle{abbrv}
\bibliography{bibliography.bib}

\end{document}